\documentclass[conference]{IEEEtran}
\IEEEoverridecommandlockouts
\usepackage{cite}
\usepackage{amsmath,amssymb,amsfonts}
\usepackage[ruled,vlined,commentsnumbered,linesnumbered,lined,boxed]{algorithm2e}
\usepackage{algorithmicx}
\usepackage{algpseudocode}
\usepackage{graphicx}
\usepackage{tabulary}
\usepackage{textcomp}
\usepackage{xcolor}
\usepackage[utf8]{inputenc}
\def\BibTeX{{\rm B\kern-.05em{\sc i\kern-.025em b}\kern-.08em
    T\kern-.1667em\lower.7ex\hbox{E}\kern-.125emX}}
\makeatletter %
\newcommand{\linebreakand}{%
  \end{@IEEEauthorhalign}
  \hfill\mbox{}\par
  \mbox{}\hfill\begin{@IEEEauthorhalign}
}
\makeatother %

\usepackage{cleveref}
\usepackage{todonotes}
\usepackage{paralist}
\usepackage[skip=0cm,list=true]{subcaption}
\begin{document}

\title{An Empirical Study of Efficiency and Privacy of Federated Learning Algorithms\\
}

\author{
\IEEEauthorblockN{Sofia Zahri}
\IEEEauthorblockA{\textit{School of EECS} \\
\textit{QMUL, London, UK}\\
s.zahri@qmul.ac.uk}
\and
\IEEEauthorblockN{Hajar Bennouri}
\IEEEauthorblockA{\textit{Collaboratory} \\
\textit{TU-Dublin, Ireland}\\
Hajar.Bennouri@tudublin.ie}
\and
\IEEEauthorblockN{Ahmed M. Abdelmoniem}
\IEEEauthorblockA{\textit{School of EECS} \\
\textit{QMUL, London, UK }\\
ahmed.sayed@qmul.ac.uk}
} 

\maketitle

\begin {abstract}

In today's world, the rapid expansion of IoT networks and the proliferation of smart devices in our daily lives, have resulted in the generation of substantial amounts of heterogeneous data. These data forms a stream which requires special handling. To handle this data effectively, advanced data processing technologies are necessary to guarantee the preservation of both privacy and efficiency. Federated learning emerged as a distributed learning method that trains models locally and aggregates them on a server to preserve data privacy. This paper showcases two illustrative scenarios that highlight the potential of federated learning (FL) as a key to delivering efficient and privacy-preserving machine learning within IoT networks. We first give the mathematical foundations for key aggregation algorithms in federated learning, i.e., FedAvg and FedProx. Then, we conduct simulations, using Flower Framework, to show the \textit{efficiency} of these algorithms by training deep neural networks on common datasets and show a comparison between the accuracy and loss metrics of FedAvg and FedProx. Then, we present the results highlighting the trade-off between maintaining privacy versus accuracy via simulations - involving the implementation of the differential privacy (DP) method - in Pytorch and Opacus ML frameworks on common FL datasets and data distributions for both FedAvg and FedProx strategies. 
\end{abstract}.

\section{Introduction}
The Internet of Things (IoT) is a leading technology that is shaping the global markets and the future of connected devices. The large amount of data exchanged daily throughout IoT networks via wireless communication technologies and the use of centralised global models for data collection, make IoT devices the prime target of privacy and security issues \cite{savazzi2020federated}. This data contains valuable information that can be leveraged for a variety of purposes, such as improving efficiency, improving decision-making processes, and enabling predictive analytics \cite{8069088,7511592,10155021}.
However, processing such huge amounts of data poses several challenges. One of the main challenges is the need for efficient data processing techniques that can handle the scale and complexity of the data generated by IoT~\cite{krishnamurthi2020overview}. 

In this context, various Machine Learning (ML) and Deep Learning (DL) models have been introduced as part of artificial intelligence (AI) to enhance the privacy and efficiency of trained models,while giving insights into the data shared between smart devices and developing advanced methods to secure IoT environments. However, these ML/DL models are being built using collected data, processed on a centralised global server \cite{RADOGLOUGRAMMATIKIS201941}.
Traditional centralised approaches may not be suitable due to limitations in bandwidth, latency, and compute resources~\cite{niknam2020federated}. In addition, data privacy becomes a major concern, as sensitive information may be exposed during centralised processing. This is where the concept of federated learning (FL) comes into play.

Federated learning flipped the paradigm by overcoming the limitations of centralised trained models \cite{McMahan2016,10.1145/3437984.3458839,10.1145/3552326.3567485}. This approach consists of training each model locally in the user's smart device instead of sending the data to the main server. The server then aggregates the model updates of the trained models via the IoT networks , to form a global model, without having access to the user's private data. This process is then repeated over multiple rounds of training \cite{McMahan2016}. The model is not trained on all the smart devices at once, instead, a sample of a fraction of them is chosen depending on the availability and reliability of these devices \cite{9060868}. This approach ensures data privacy and reduces the need for data transfer, which helps address the privacy and efficiency challenges highlighted in the problem.

 In our initial simulation, we conducted a thorough comparison of the accuracy and loss metrics of two distinct federated learning algorithms, namely FedAvg and FedProx \cite{li2020federated}. To ensure optimal accuracy and precision, we utilised the innovative Flower framework \cite{beutel2020flower}, The Flower framework is compatible with various client environments and can handle many clients, including mobile and embedded devices. It allows effective analysis and comparison of algorithm performance under controlled conditions based on different parameters such as the proximal term values and straggler values. Our results provide valuable insights into the efficiency of the FedProx approach considering the heterogeneity of data in IoT networks. In the second scenario, we used Opacus with Pytorch to implement differential privacy (DP) for both the FedAvg and FedProx \cite{dwork2008differential}. Then, we conducted experiments and presented the results of these two deferentially private approaches, namely DP\_FedAvg and DP\_FedProx.Hence,we concluded that by implementing DP,we can enhance the privacy of these strategies, but there is a clear trade-off between privacy preservation and model accuracy.

 Our contributions to this work are as follows:
 
 \begin{enumerate}
     \item We emphasise the emerging dominance of federated learning as the key data analytic method in IoT environments.
     \item We study and analyse mathematically and empirically key federated learning algorithms and compare them from two main aspects, i.e., efficiency and privacy.
     \item We present our experimental results using mainstream federated learning framework, flower, and show on various benchmarks that the choice of the federated learning or privacy-preserving algorithm has a noticeable impact on the model quality and convergence.
 \end{enumerate}

 This paper is organised as follows. We first present background on federated learning and its emerging role and advantages in IoT environments in \cref{sec: background}. Then, we present the mathematical analysis of different federated learning algorithms in \cref{sec:math}. Then, we discuss the experimental details and results for both scenarios in \cref{sec:expreriments}. We end the paper with a summary and a discussion in the \cref{sect:concl}.

\section{Federated Learning}
\label{sec: background}

\subsection{Federated Learning model}

 Federated learning (FL) techniques are used to manage the consensus learning process in a decentralised manner. In this approach, a central server is responsible for coordinating the global learning objective, while multiple devices train the model using locally collected data. The FL iterative process starts by creating an initial model at the central server and then sending it to the Edge Machine Learning Operations (MLOps) of IoT smart devices \cite{9787703}.
 
MLOps is composed of three stages \cite{9610376} :

\begin{enumerate}
    \item \textbf{ML Design:} which consists of data sourcing and availability, data labelling, and data versioning.
   \item \textbf{Model development:} which consists of model architecture and engineering, model training, and model testing and validation.
    \item \textbf{Operations:} which consists of Model versioning, Model deployment then Model prediction.
\end{enumerate}

\begin{figure}[t!]
\centering
\includegraphics[width=\columnwidth]{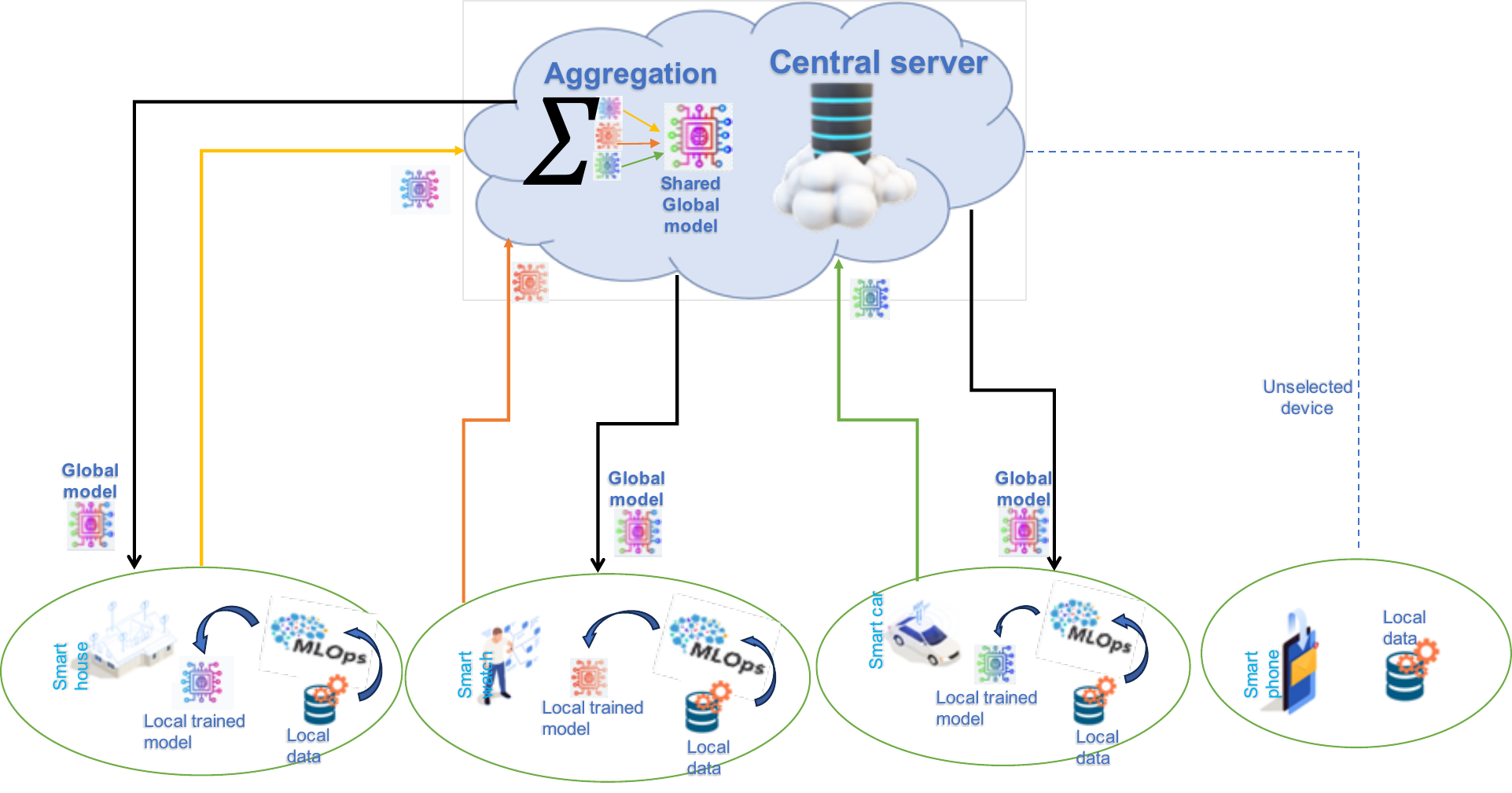}
\caption{FL architecture - Model Aggregation.}
\label{fig1}
\end{figure}

Each smart device in the IoT network receives a pre-trained or non-trained initial ML model and then trains it locally with the use of its local data. The smart devices then send the locally trained models back to the central server. The central server then aggregates the local models to build a shared global model as shown in \cref{fig1}. The shared global model is then sent back to a sub-sample of the smart devices for evaluation. The model updates are then evaluated to determine the accuracy and identify any improvements over the previously sent version. The aggregation and sharing of the updated global model is repeated until the global model reaches a satisfactory level of accuracy or until a predetermined number of iterations has been reached.\\

\begin{figure}[t!]
\centering
\includegraphics[width=0.7\columnwidth]{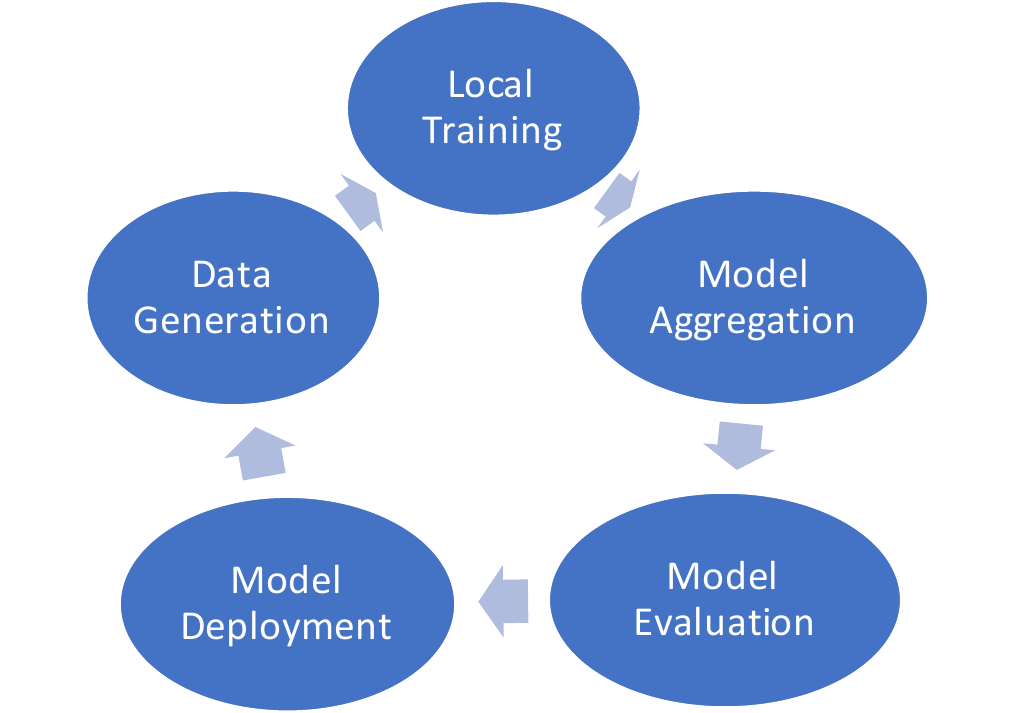}
\caption{Federated Learning iterative process.}
\label{fig2}
\end{figure}

The life cycle of the FL process is presented in \cref{fig2}. It starts with the data generation on the clients and ends with the model deployment by the server. First, after the clients owning the data join the process, they receive the current version of the global model from the server. Then, the local training is invoked by the server on the clients. After sharing the newly produced local models with the server, the model aggregation occurs on the server followed by the model evaluation on the test dataset or clients. Finally, the server deploys the new global model on the target devices. This process repeats throughout the model's lifetime. It is worth noting that after the creation of a new model on the server, the evaluation phase will determine if the globally updated model received in each round achieves the recommended expectations or if it's time to start the process once again.

\subsection{Advantages of Federated learning}

In this section, we will explain the advantages of using federated learning as a decentralised approach and summarise them in the following \cref{tab1}.

\begin{table}[h!]
\caption{Advantages of Using Federated Learning\label{tab1}}
\begin{center}
  \begin{tabulary}{0.7\textwidth}{ | p{2cm} | p{5.5cm} |}
  \hline
    \ \textbf{Advantages} & \textbf{Explanation}                                                \\
    \hline
     Privacy Preservation & Federated learning allow models to be trained directly on user devices, keeping sensitive data decentralised. This reduces the risk of a data or privacy breach.\\
    \hline
    Reduced Communication	 & Instead of moving data to a centralised cloud server, only model updates are transmitted, reducing the amount of communication needed, bandwidth required, and latency. \\
    \hline
    Improved Model Accuracy &  By leveraging diverse data from multiple devices or users, federated learning helps build more accurate models through greater diversity and inclusiveness in training.\\
    \hline
  Faster Model Training	& Using federated learning, models can be trained in parallel on multiple devices, enabling faster updates and reducing total training time.   \\
    \hline
  Low Latency during Inference	 & 	Now that models are trained and run locally on users' devices, this helps reduce latency when inferring, providing fast, real-time predictions without depending on a constant network connection.  \\
    \hline
	Maintaining full control over Data & Federated learning ensures users have full control over their data, as it remains stored on their own devices. This reduces the risk of privacy breaches and keeps data ownership in the hands of users. \\

\hline
	Improved Energy Efficiency of Devices & By training models locally, federated learning reduces reliance on cloud servers, resulting in more efficient use of local device energy resources, contributing to better overall energy efficiency.\\
	\hline
\end{tabulary}  
\end{center}
\end{table}
Table \ref{tab1} highlights the benefits of using federated learning in terms of preserving data privacy, reducing data traffic, improving model accuracy, speeding up training, low latency during inference, data mastery by users, and improved energy efficiency. These benefits result from training models on user devices, selectively sharing model updates, and decentralising sensitive data.

\subsection{Challenges of Federated learning}

When it comes to federated learning in IoT networks, there are several challenges and factors to consider, including restricted resources, computational demand, high cost, and the diversity of data and devices as mentioned by the authors \cite{10.1145/3517207.3526969,10061708}.

\section{Mathematics Analysis}
\label{sec:math}

While there has been significant efforts to improve network performance and latency~\cite{7511372,7511592,8885278,8647878,9372846,IPCCC} and workload management~\cite{8885122,3404447}, the FL framework is a valuable tool for reducing communication overhead in IoT networks. This means that instead of transferring raw data, only periodic model updates are sent to the orchestrator server. Additionally, IoT systems can generate vast amounts of data that can vary based on the hardware capabilities of each device. The variation in the mixed data can have a significant impact on federated networks. For this reason, Google has introduced federated averaging (FedAvg) as a key optimisation algorithm for federated systems to tackle the issues arising from system heterogeneity~\cite{McMahan2016}.

\subsection{FedSGD vs FedAvg}

FedAvg is based on the FL training algorithm on Stochastic Gradient Descent (SGD)\cite{45187}. Federated SGD (FedSGD) involves calculating a single batch gradient during a single round of communication, using a randomly selected client.

At the beginning of each communication round t, the server selects a random fraction $C$ of clients for performing computational model training. \\
 Considering a training dataset containing $n$ pairs of (input-output) $(x_i ,y_i)$ samples with $ i \in [1,n]$ :
 
 The goal in traditional deep learning model training is to find a set of parameters $w$, that maximises the probability of the outputting $y_i$, given the input $x_i$, meaning the average of the probability $p(y_i)$ is maximised such as the following \cref{eq:1} is maximised : 
 \begin{equation}
     \frac{1}{n} \sum_{i=1}^{n} p(y_i|x_i,w)
     \label{eq:1}
\end{equation}

 Which is equivalent to minimising \cref{eq:2} :
 
 \begin{equation}
       \frac{1}{n} \sum_{i=1}^{n} -log (p(y_i|x_i,w))
       \label{eq:2}
\end{equation}
In deep learning, we generally take: 
$ f_i(w)= l(x_i,y_i,w) $, which is, the loss function of the prediction on chosen samples $(x_i,y_i) $ made with model parameters $w$, such as   $l(x_i,y_i,w) = -log (p(y_i|x_i,w))$.
\\

From \cref{eq:2}, we conclude that the model training objective is defined as follows:

\begin{equation}
\min_{w\in R^d} f(w) \hspace{0.5cm} where \hspace{0.5cm} f(w) = \frac{1}{n}\sum_{i=1}^{n}f_i(w)
\label{eq:3}
\end{equation}

We assume that we have distributed smart devices of $k$ clients. where $p_k$ is the relative impact of each device or set of data points on $k$th client or device, with two settings: 
\begin{itemize}
    \item $p_k =\frac{n_k}{n}$  or  $p_k= \frac{1}{n}$ .
    \item  $n=\sum_{i}n_k$  is the total number of the chosen samples.

\end{itemize}
The \cref{eq:3} becomes  as follows :

\begin{equation}
f(w) = \sum_{k=1}^{K}\frac{n_k}{n}F_k (w) \hspace{0.2cm} where \hspace{0.2cm} F_k(w) = \frac{1}{n_k}\sum_{i \in P_k} f_i (w)
\label{eq:4}
\end{equation}   
As stated earlier, the main goal is to minimise the objective function as follows : 

\begin{equation}
\min_{w}F(w)\hspace{0.3cm}where\hspace{0.3cm} F(w) = \sum_{k=1}^{H}p_kF_k(w)
\end{equation}

The last equation represents the final model function with the following variables : 
\begin{itemize}
    \item $F (w)$ is the global objective function.
    \item $F_k (w)$ is the local objective function for the $k$th client or device.
    \item $H$ is the total number of clients or devices.
    \item  $p_k \geq 0$ and $\sum_k p_k = 1$.\\     
\end{itemize}

In Federated Learning, choosing a client with $n_k$ training samples represents a random selection, similar to traditional deep learning. Therefore, in FedSGD, the fraction of clients chosen during each round $t$, known as the $C$ fraction, is typically set to  1 to regulate the overall batch size. \\
In a single round $t$ : 
\begin{enumerate}
    \item The central server sends the initial model with parameters $w_t$ to each client. 
    \item Each client $k$ computes the gradient $g_k$ by using its local data and by applying the following formula $g_k = \nabla F_k (w_t)$.
    \item Then each client $k$ submits the $g_k$ to the central server.
    \item    The gradients aggregation takes place at the central server.
\end{enumerate}

In order to generate a new model the aggregation of the updates is achieved by $w_t+1 \leftarrow w_t - \eta \nabla f(w_t)$ and then applying \cref{eq:4} to obtain $w_t+1 \leftarrow w_t - \eta \sum_{k=1}^{K} \frac{n_k}{n}g_k$, where, 
\begin{inparaenum}
    \item $\eta$ is the learning rate;
    \item $n$ the total samples;
    \item $K$ Clients; 
    \item $n_k$ the sample on client $k$; and
    \item Client fraction $C =1$.
\end{inparaenum}

Concerning FedAvg, the central server takes the weighted average of the resulting models after performing more than one batch of updates on the client's local data.

\begin{algorithm}
\caption{Federated Averaging (FedAvg)}
\label{alg:cap}
Server initialises the model parameters $w_0$\;

On Server:
\For{each round $t = 1,2,3,4,....$}
{
  $S_t \leftarrow$  Select random sample of $H$ clients\;
  \For{each client $k \in S_t$ in parallel}
  {
    
	$w_{t+1}^{k} \leftarrow$ ParameterUpdate ($K, w_t$)\; 
	$w_t+1 \leftarrow \sum_{k=1}^{K} w_{t+1}^{k}$\;
  }
}
On Clients run ParameterUpdate($k,w$) for client $k$:
$B \leftarrow$ Split $P_k$ into local minibatches of size $B$\;
\For{each local epoch $i$ from $1$ to $E$} 
{ 
	\tcc{$E$ is number of local epochs}
	\For{batch $b \in$ $B$}
	{
		$w \leftarrow w - \eta \nabla l(w ; b)$\;
	}
}
Send $w$ to server\;

\end{algorithm}
 
\subsection{FedProx}

Federated learning is a popular method for training deep learning models in Internet of Things (IoT) networks. However, the non-iid data distribution among different clients can lead to significant challenges in achieving optimal performance. To address this issue, a novel optimisation technique called FedProx has been proposed.
FedProx represents an innovative upgrade of the FedAvg strategy for federated learning, which offers a reliable means of achieving convergence. This approach ensures successful learning outcomes, even when dealing with data that is not independently or identically distributed (non-IID), as well as different system characteristics.\\

Authors in \cite{li2020federated} present FedProx, a framework that tackles two main challenges in federated learning which are:
\begin{itemize}
    \item \textbf{Systems heterogeneity:} With FedAvg, participating devices cannot adjust their workload based on their system constraints. Instead, devices that fail to complete a certain number of epochs within a specific time frame are often removed from the model training process \cite{47976}.
    \item \textbf{Statistical heterogeneity:} Empirical evidence suggests that FedAvg diverges in situations where the data is not identically distributed among client devices \cite{McMahan2016}.
\end{itemize}

FedProx introduces a new regularisation term, referred to as the "proximal term" which helps to mitigate the impact of non-iid data distribution. By incorporating this term into the standard federated optimisation objective, FedProx enhances the overall efficiency of the model, making it a promising technique for optimising deep learning models in federated settings.\\

The proximal term is defined by the same authors 
\cite{li2020federated } is simply the regularisation term in the local model used on each client. In the following equation (5), we present the minimisation of the local objective function in round $ t+1$, while we note the proximal term as being :
\begin{math} \frac{\mu}{2} \| w - w^t \|^2 \end{math}.
   
\begin{equation}
\label{eq:5}
    \min_{w} h_k(w;w^t) = F_k(w) + \frac{\mu}{2} \| w - w^t \|^2
\end{equation}.
In \cref{eq:5}, we have the following variables: 
\begin{itemize}
    \item $F_k(w)$ as the local loss function for the $k$-th client that is supposed to be minimised.
    \item $w$ are the local hyper parameters.
    \item $w^t$ are the global weights of the server at epoch $t$.
    \item $h_k(w;w^t)$ represents the objective function that the $k$th client is trying to minimise.

\end{itemize}

\section{Experimental Results}
\label{sec:expreriments}

\subsection{Experimental Setup}
In this paper, the experimental setup will focus on:

\begin{enumerate}
    \item Comparison of convergence analysis to evaluate aggregation strategies in a federated learning framework on Non-IID Data. \cite{9716797,fedavg}.
    \item Insights into the differential privacy in federated learning techniques and algorithms tailored for IoT networks.\\
\end{enumerate}

The following \cref{tab2} outlines the hyper parameters necessary for conducting our first simulation \cref{FL efficiency using Flower} using the flower framework and poetry environment in Python. These settings are carefully considered for model training using heterogeneous non-IID, unbalanced MNIST datasets.

\begin{table}[h]
\caption{Model Training hyper parameters \label{tab2}}
\begin{center}
  \begin{tabulary}{0.5\textwidth}{ | p{2.5cm} | p{4.5cm} |}
  \hline
    \ \textbf{Description} & \textbf{Configuration Values}                                                \\
    \hline
     Total clients & 1000 \\
     \hline 
     Clients per round & 10 \\
     \hline
     Number of rounds & 100 \\
     \hline
     Server device & GPU \\
     \hline
     Data partition & Pathological with power law \\
     \hline
     partition settings & 2 Classes per client \\
     \hline
     Optimiser & SGD with proximal term \\
     \hline
     Dataset configuration &  Non-IID data - Unbalanced data - Law power\\
     \hline
     Fitures configuration & Drop client: false \\
     \hline
     Model target & Fedprox models using logistic regression \\
     \hline
      Number of classes & 10 \\
     \hline
     Strategy target &  Flower server with FedProx strategy\\
     \hline
     
\end{tabulary}  
\end{center}
\end{table}  

\subsection{FL efficiency using Flower }
\label{FL efficiency using Flower}

The Flower framework is a new approach to end-to-end federated learning that allows for a smoother transition from experimental simulation research to system research on real-edge devices in IoT networks \cite{beutel2020flower}. The simulation implements the following two models:
 \begin{itemize}
\item A logistic regression model (LR) used in the FedProx work for MNIST dataset \cite{li2020federated}.
\item  A two-layer convolution neural network (CNN) similar to the FedAvg work \cite{McMahan2016}.
 \end{itemize}
 
\begin{figure}[!t]
    \centering
    \includegraphics[width=0.75\linewidth]{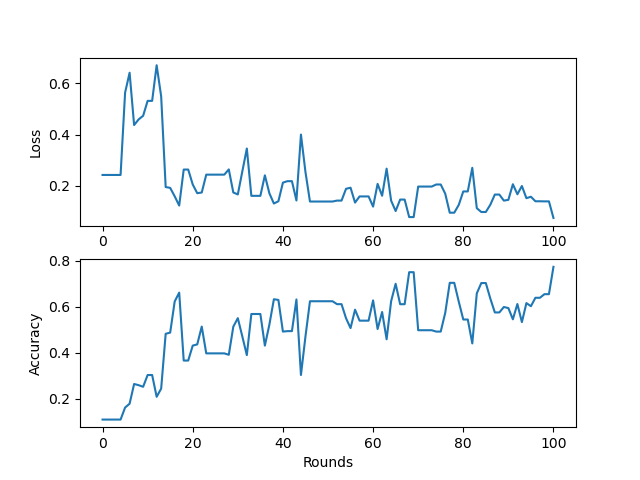}
    \centering\caption{FedAvg Loss/Accuracy with 90\% straggler \& $\mu=0.0$. }
    \label{fig3}
\end{figure}

\begin{figure}[!t]
    \centering
    \includegraphics[width=0.75 \linewidth]{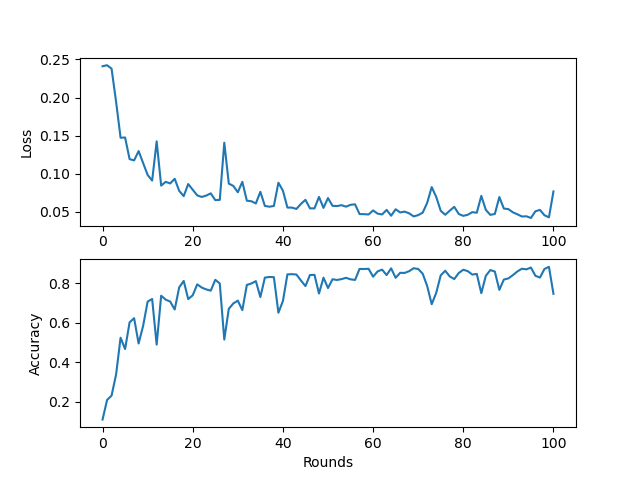}
    \centering\caption{FedProx Loss/Accuracy with 90\% straggler \& $\mu=1.0$. }
    \label{fig4}
\end{figure}

The following \cref{fig3}  and \cref{fig4} represent the results of the implementation of flower baselines following the experiments of the paper \cite{li2020federated}. We used the dataset MNIST from PyTorch's Torchvision. The task of the experiment is to conduct image classification. It draws the comparison of  FedProx accuracy-loss convergence on Non-IID Data with FedAvg accuracy-loss convergence.

The MNIST dataset is divided into 1000 clients using a pathological split. Each client has examples of two out of ten class labels. The number of epochs is 10, the batch size is 10. For the training hyperparameters, the number of clients per round is 10 and the number of rounds is 100. The learning rate is set at 0.03 and the stragglers' fraction at 0.9 meaning 90 \% stragglers and proximal $\mu$ is set to 0.0 for FedAvg \cref{fig3} and proximal $\mu$ is set to 1 for FedProx \cref{fig4}.

The environment setup consists of the installation and activation of the installation and activation of Poetry environment using pip in Python and the installation of Pytorch with Cuda version 11.6. During the first FedAvg simulation, a variation is used to drop the clients that were flagged as stragglers.

Our model training steps while running the first FedAvg simulation are as follows:
 \begin{itemize}
     \item Dataset partitioning configuration: ('iid': False, 'balance': False, 'power\_law': True).
     \item Starting Flower simulation, config: ServerConfig : (num\_rounds$=100$, round timeout $=None$).
     \item Flower VCE: Ray initialised with resources ('object\_store\_memory': 40004159078.0, 'memory': 83343037850.0, 'node:138.37.88.24': 1.0, 'accelerator type: P100': 1.0, 'GPU': 2.0, 'CPU': 16.0).
 \end{itemize}

\begin{figure*}[!t]
    \centering
    \includegraphics[width=0.8\linewidth]{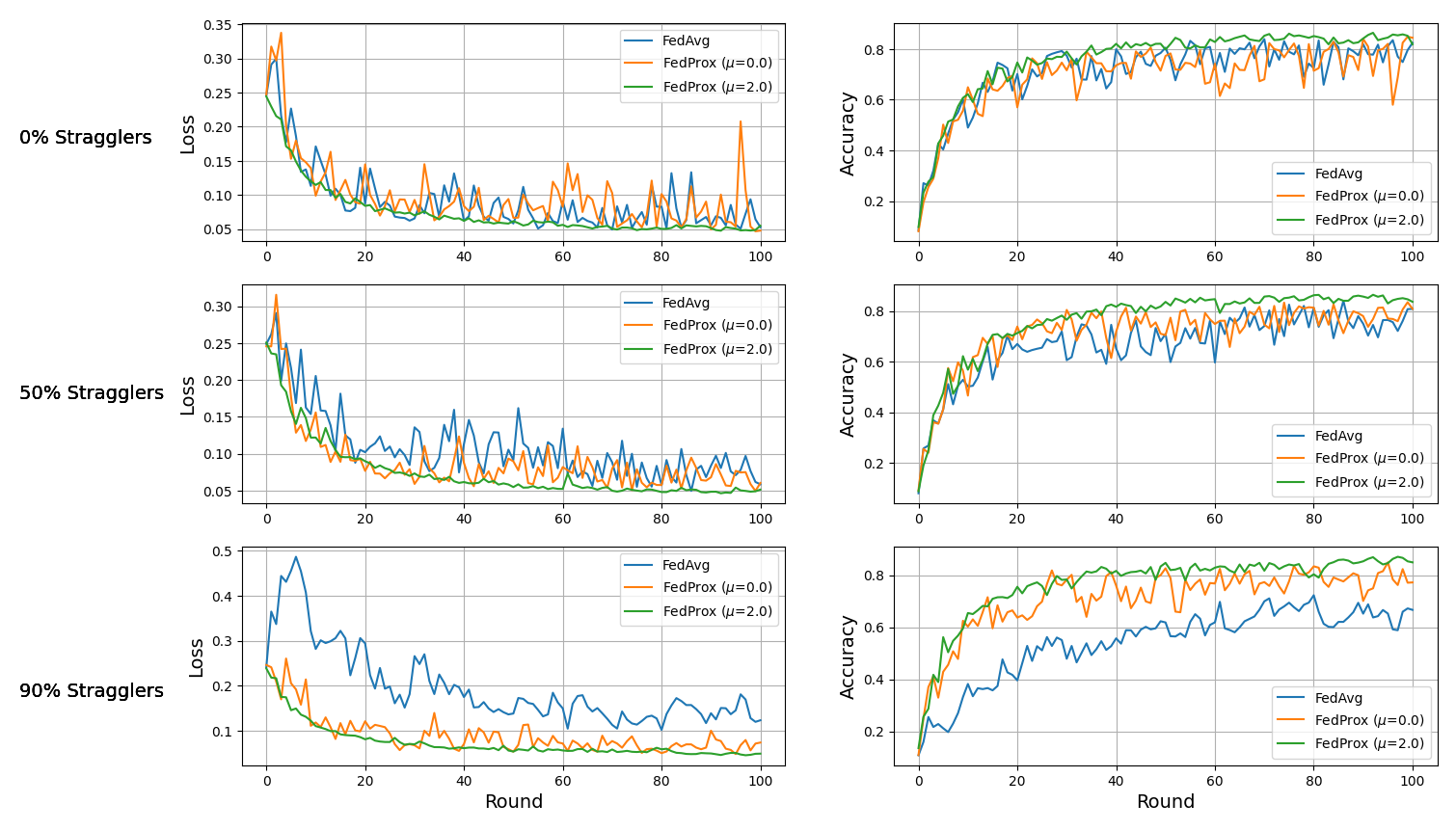}
    \caption {FedProx vs FedAvg Accuracy \& Loss with 0 \% , 50\% and 90\% stragglers and two values of proximal $\mu$ = (0.0, 2.0).}
    \label{fig5}
\end{figure*}

In the following \cref{fig5}, we present the results for both FedProx and FedAvg with different configurations multiple times while iterating through different values of the proximal term $\mu$ and the percentage of stragglers as follows:
\begin{itemize}
    \item Proximal term ($\mu$): $\mu$=0.0 and $\mu$=2.0.
    \item Stragglers Percentage: 0\%, 50\% and 90\% Stragglers.
\end{itemize}

Using FedProx instead of FedAvg can result in significant improvements in convergence, especially in scenarios with varying device capabilities. To evaluate the impact of system heterogeneity, we conducted simulations where 0\%, 50\%, and 90\% of devices were designated as stragglers and dropped by FedAvg \cite{li2020federated}. 
Through a comparison of FedAvg and FedProx, it is apparent that incorporating varying workloads can improve convergence, especially in heterogeneous smart device environments. This observation is particularly relevant when considering the value of $\mu$ set to 0.

When comparing the performance of FedProx with a proximal term value of 0 to that of FedProx with a value greater than 0 or equal to 2, it becomes evident that increasing the proximal term value offers significant benefits. FedProx with a proximal term value of 2 exhibits more stable convergence and enables convergence in scenarios where methods would otherwise diverge, including situations with heterogeneity (50\% and 90\% stragglers) or without heterogeneity (0\% stragglers). These findings suggest that a higher proximal term value is advantageous for optimizing the performance of FedProx. In cases where there are no stragglers and the value of the parameter $\mu$ is 0, the FedProx algorithm becomes equivalent to the FedAvg algorithm.

Based on the simulation results, it has been concluded that FedProx provides better and more consistent convergence compared to FedAvg. This is especially apparent in situations where there is a high level of heterogeneity, as FedProx exhibits significantly more stable and accurate convergence behaviour compared to FedAvg.

\subsection{FL Differential Privacy (DP) with Opacus}

While federated learning is a step towards privacy, it is not enough to fully achieve it. During the training process, privacy remains a serious concern as communicating model updates, e.g., gradient information can potentially reveal sensitive information. This is while sharing the model weights or updates instead of the raw data, can reveal sensitive information, either to the central server or to a third party \cite{aledhari2020federated}.

Recent research has suggested various methods to address this issue. One such solution involves encrypting the model's weights before transmitting them to the central server, which requires more computing power. Another approach is to use differential privacy (DP)\cite{8049725}. DP operates on the concept of adjacent data, which ensures that the model's performance remains unaffected when a specific element is added or removed from the database \cite{dwork2008differential}.

Opacus, the Pytorch-based tool, enables flexible training of models using differential privacy (DP) with minimal impact on client-side code changes and training performance \cite{2021arXiv210912298Y}. The PrivacyEngine abstraction offered by Opacus facilitates the tracking of the privacy budget - a crucial mathematical concept in DP - and works on the model's gradients. Opacus is an effective tool for training DP based models. It uses a stochastic gradient descent (SGD) optimiser that applies noise to the gradients during each step of the descent, rather than to the data or to the model parameters.

In the experiments, we apply differential privacy (DP) to both FedAvg and FedProx \cite{8049725}. We focused on non-IID, MNIST dataset and used Pytorch's Opacus library for this purpose. To achieve a differentially private FL model, we adhered to the standard practice of introducing an independent noise to the gradients in each training iteration. Specifically, we incorporated Gaussian noise within both FedAvg and FedProx to preserve the privacy of each sample while limiting the impact on the accuracy of the global model.

\subsubsection{DP\_FedAvg}
\begin{figure}[!t]
    \centering
    \includegraphics[width=0.75\linewidth]{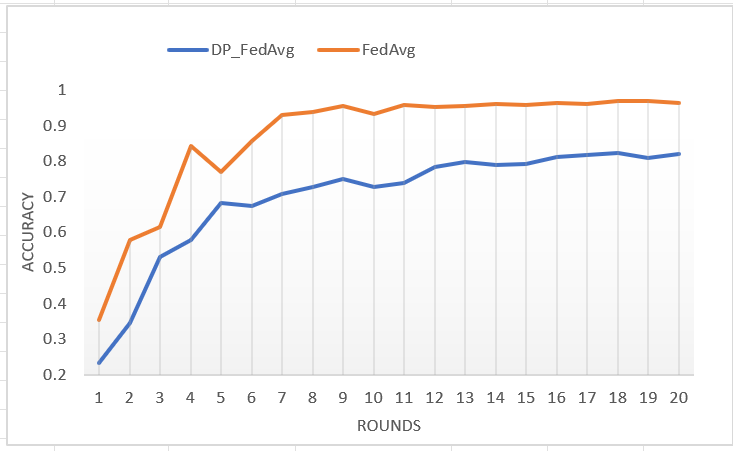}
    \caption{FedAvg Accuracy using differential privacy}
    \label{fig6}
\end{figure}

In \cref{fig6} we compared FedAvg without DP and FedAvg with DP, in order to understand the impact that differential privacy can have on the model accuracy's performance. We clearly see the cost of adding a DP to enhance the privacy of the FedAvg algorithm. This difference in the performance goes back to the core concept of differential privacy, consisting of adding (unbiased) noise to the parameter gradients that use the model to update its weights. In each iteration, we prevent the model from memorising its training path while still allowing the learning process to proceed.

\subsubsection{DP\_FedProx}

\begin{figure}[!t]
    \centering
    \includegraphics[width=0.75\linewidth]{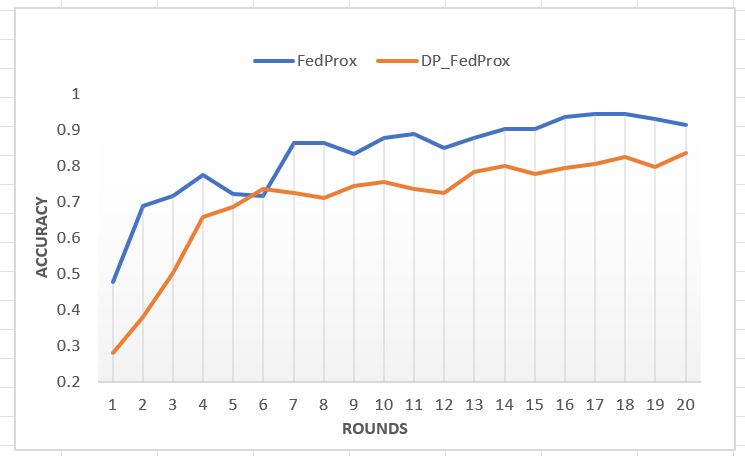}
    \caption{FedAprox Accuracy using differential privacy}
    \label{fig7}
\end{figure}

Considering the use of the FedProx algorithm in IoT networks, we added the differential privacy (DP) to FedProx, in order to measure the model performance on heterogeneous data distributions. The model accuracy performance in \cref{fig7} increases with each round and attends higher accuracy values up to 0.9146 in the 20th round without DP. In case of FedProx with DP, the model accuracy reaches the value of 0.8353 in the 20th round while preserving its privacy.

\section{Conclusion and Future Work}
\label{sect:concl}

In this paper, we focused on introducing the advantages of using federated learning for IoT networks. We have briefly touched on the challenges that come with the use of FL such as high computation demands which means expensive communication in federated networks, systems heterogeneity which differs from one device to another and one environment to another, statistical heterogeneity related to the type of computed data. We have proposed two scenarios covering the efficiency and privacy in IoT networks using FedAvg and FedProx. The fundamental principle of FedProx involves the correlation between statistical heterogeneity and the heterogeneity of systems. The elimination of straggling devices from a network as a result of systems constraints may inadvertently intensify statistical heterogeneity. In order to address this issue, FedProx modifies the FedAvg method by allowing partial work to be done on devices based on their underlying systems constraints, while incorporating the proximal term to integrate the partial work safely. From a reparameterisation perspective, adjusting the proximal term in FedProx is equivalent to changing the number of local epochs in FedAvg.Hence, the first simulation confirms the efficiency of FedProx in IoT networks using the flower framework. The second simulation was designed to focus on the implementation of differential privacy (DP) through the utilisation of Opacus. The aim of this was to ensure the privacy of each model training sample while minimising the impact on the accuracy of the global model. Opacus is used to integrate DP into the model with ease. The simulation results demonstrated that DP\_FedAvg and DP\_FedProx can achieve high accuracy using DP-SGD in Opacus, thereby enabling privacy-preservation of the model.

\bibliographystyle{ieeetr}
\bibliography{refs}
\end{document}